\documentclass[10pt,twocolumn,letterpaper]{article}

\usepackage{iccv}
\usepackage{times}
\usepackage{epsfig}
\usepackage{graphicx}
\usepackage{amsmath}
\usepackage{amssymb}

\usepackage{algorithm,algorithmic}
\usepackage{amsmath}
\usepackage{amsfonts} 
\usepackage{booktabs}
\usepackage{url}
\usepackage{cite}
\usepackage{microtype}
\usepackage{multirow}

\usepackage{color}
\usepackage{bbm}
\newcommand\num[1]{\textcolor{black}{{#1}}}

\newif\ifpaper
\newif\ifsupplementary
\papertrue
\supplementarytrue
\newcommand\multiplier[0]{108}
\newcommand\numlabels[0]{10}
\newcommand\transferdiff[0]{5.16} % was 8.27
\newcommand\rsquare[0]{0.778}

\usepackage{verbatim}
\newcommand{\object}[1]{\texttt{#1}}
\newcommand{\predicate}[1]{\texttt{#1}}
\newcommand{\relationship}[3]{$<$\texttt{#1} - \texttt{#2} - \texttt{#3}$>$}

\usepackage[pagebackref=true,breaklinks=true,letterpaper=true,colorlinks,bookmarks=false]{hyperref}

\iccvfinalcopy % *** Uncomment this line for the final submission

\def\iccvPaperID{3739} % *** Enter the ICCV Paper ID here

\ificcvfinal\fi
\begin{document}

%%%%%%%%% TITLE
\title{Scene Graph Prediction with Limited Labels}

\author{Vincent S. Chen, Paroma Varma, Ranjay Krishna, Michael Bernstein, Christopher R\'e, Li Fei-Fei\\
Stanford University\\
{\tt\small \{vincentsc, paroma, ranjaykrishna, msb, chrismre, feifeili\}@cs.stanford.edu}
}

\ifpaper
    \maketitle

    %%%%%%%%% ABSTRACT
    \begin{abstract}
       Visual knowledge bases such as Visual Genome power numerous applications in computer vision, including visual question answering and captioning, but suffer from sparse, incomplete relationships. All scene graph models to date are limited to training on a small set of visual relationships that have thousands of training labels each. Hiring human annotators is expensive, and using textual knowledge base completion methods are incompatible with visual data. In this paper, we introduce a semi-supervised method that assigns probabilistic relationship labels to a large number of unlabeled images using few labeled examples. We analyze visual relationships to suggest two types of image-agnostic features that are used to generate noisy heuristics, whose outputs are aggregated using a factor graph-based generative model. With as few as $\numlabels$ labeled examples per relationship, the generative model creates enough training data to train any existing state-of-the-art scene graph model. We demonstrate that our method outperforms all baseline approaches on scene graph prediction by $\num{\transferdiff}$ recall@100 for \textsc{PREDCLS}. In our limited label setting, we define a complexity metric for relationships that serves as an indicator (\num{$R^2 = \rsquare$}) for conditions under which our method succeeds over transfer learning, the de-facto approach for training with limited labels.

    \end{abstract}
    
    %%%%%%%%% BODY TEXT
    \section{Introduction}
    In an effort to formalize a structured representation for images, Visual Genome~\cite{krishnavisualgenome} defined \textbf{scene graphs}, a formalization similar to those widely used to represent knowledge bases~\cite{guodong2005exploring,culotta2004dependency,zhou2007tree}. 
Scene graphs encode objects (e.g. \object{person}, \object{bike}) as nodes connected via pairwise relationships (e.g., \predicate{riding}) as edges. This formalization has led to state-of-the-art models in image captioning~\cite{anderson2016spice}, image retrieval~\cite{johnson2015image,schuster2015generating}, visual question answering~\cite{johnson2017inferring}, relationship modeling~\cite{krishna2018referring} and image generation~\cite{johnson2018image}.
However, all existing scene graph models ignore more than $98\%$ of relationship categories that do not have sufficient labeled instances (see Figure~\ref{fig:sparsity}) and instead focus on modeling the few relationships that have thousands of labels~\cite{xu2017scene,lu2016visual,zellers2017neural}.

Hiring more human workers is an ineffective solution to labeling relationships because image annotation is so tedious that seemingly obvious labels are left unannotated.
To complement human annotators, traditional text-based knowledge completion tasks have leveraged numerous semi-supervised or distant supervision approaches~\cite{bordes2013translating,bordes2014semantic,gardner2014incorporating,nickel2011three}. 
These methods find syntactical or lexical patterns from a small labeled set to extract missing relationships from a large unlabeled set. 
In text, pattern-based methods are successful, as relationships in text are usually \textbf{document-agnostic} (e.g. \relationship{Tokyo}{is capital of}{Japan}). 
Visual relationships are often incidental: they depend on the contents of the particular image they appear in. 
Therefore, methods that rely on external knowledge or on patterns over concepts (e.g. most instances of \object{dog} next to \object{frisbee} are \predicate{playing with} it) do not generalize well. The inability to utilize the progress in text-based methods necessitates specialized methods for visual knowledge.

\begin{figure}[t]
  \centering
    \includegraphics[width=\linewidth]{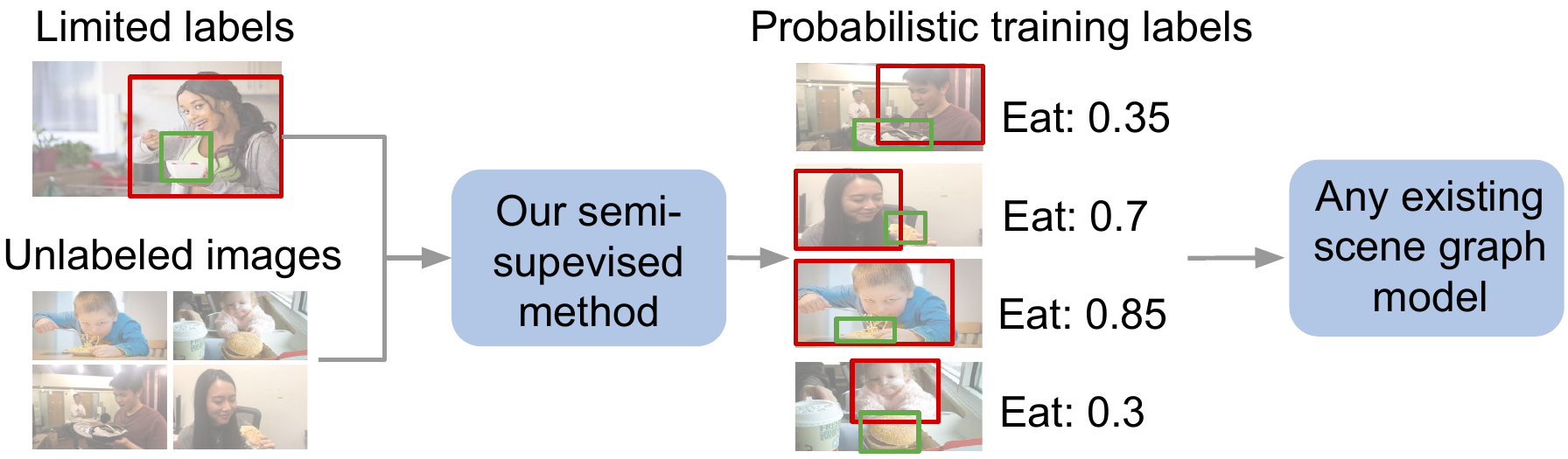}
  \caption{Our semi-supervised method automatically generates probabilistic relationship labels to train any scene graph model.}
  \label{fig:pull_figure}
  \vspace{-1em}
\end{figure}

\begin{table*}[t]
\centering
{\small
    \begin{tabular}{@{}lcccccccccc@{}}
    \sc{Num. Labeled ($\leq n$)} & \textbf{200} & \textbf{175} & \textbf{150} & \textbf{125} & \textbf{100} & \textbf{75} & \textbf{50} & \textbf{25} & \textbf{10} & \textbf{5} \\
    \toprule
    \sc{\% Relationships}   & 99.09        & 99.00        & 98.87        & 98.74        & 98.52        & 98.15       & 97.57       & 96.09       & 92.26       & 87.28     
\end{tabular}
}
\vspace{-1em}
\end{table*}

\begin{figure*}[t]
  \centering
    \includegraphics[width=0.8\linewidth]{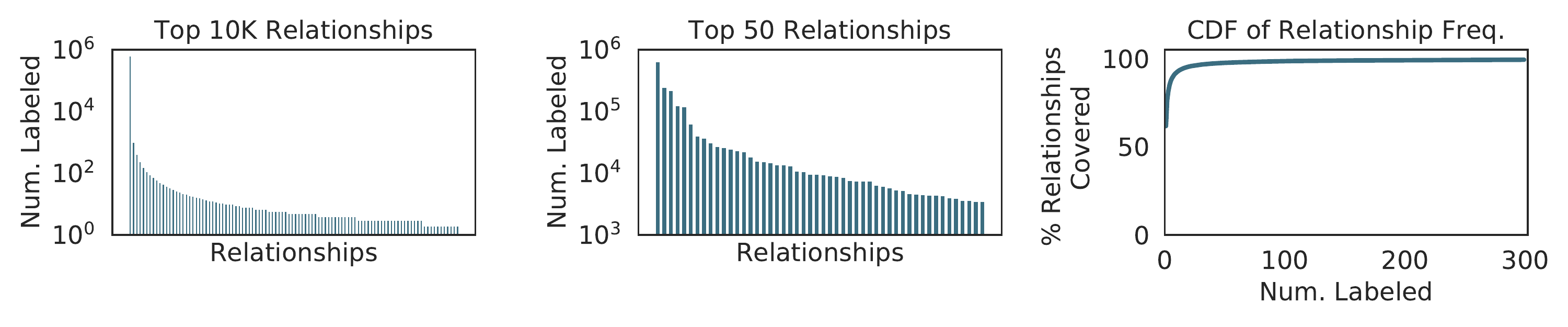}
  \caption{Visual relationships have a long tail (left) of infrequent relationships. Current models \cite{zellers2017neural,xu2017scene} only focus on the top $50$ relationships (middle) in the Visual Genome dataset, which all have thousands of labeled instances. This ignores more than $98\%$ of the relationships with few labeled instances (right, top/table).}
  \label{fig:sparsity}
  \vspace{-1.0em}
\end{figure*}

\begin{figure*}[t]
  \centering
    \includegraphics[width=\linewidth]{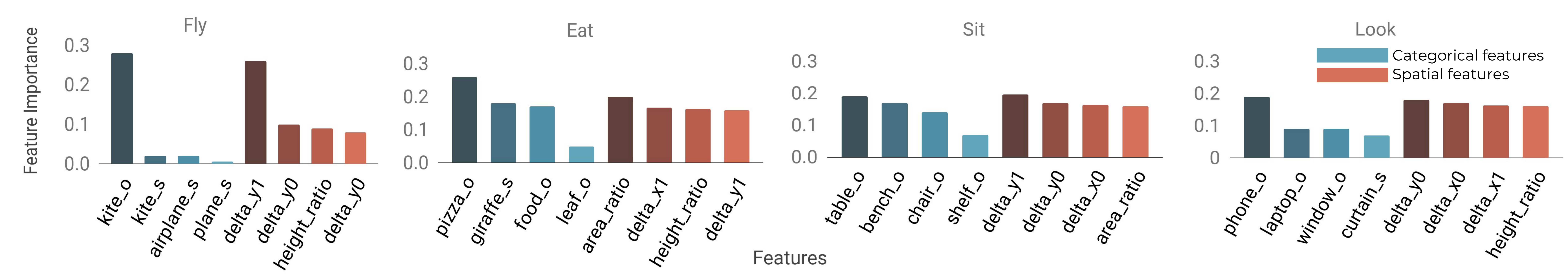}
  \caption{Relationships, such as \predicate{fly}, \predicate{eat}, and \predicate{sit} can be characterized effectively by their categorical (\texttt{s} and \texttt{o} refer to subject and object, respectively) or spatial features. Some relationships like \predicate{fly} rely heavily only on a few features --- \object{kites} are often seen high up in the sky.}
  \label{fig:importance}
\end{figure*}

In this paper, we automatically generate missing relationships labels using a small, labeled dataset and use these generated labels to train downstream scene graph models (see Figure~\ref{fig:pull_figure}).
We begin by exploring how to define \textbf{image-agnostic} features for relationships so they follow patterns across images. 
For example, \predicate{eat} usually consists of one object consuming another object smaller than itself, whereas \predicate{look} often consists of common objects: \object{phone}, \object{laptop}, or \object{window} (see Figure~\ref{fig:importance}). 
These rules are not dependent on raw pixel values; they can be derived from image-agnostic features like object categories and relative spatial positions between objects in a relationship. 
While such rules are simple, their capacity to provide supervision for unannotated relationships has been unexplored. 
While image-agnostic features can characterize \textit{some} visual relationships very well, they might fail to capture \textit{complex relationships} with high variance. To quantify the efficacy of our image-agnostic features, we define ``subtypes'' that measure spatial and categorical complexity (Section~\ref{sec:trends}). 

Based on our analysis, we propose a semi-supervised approach that leverages image-agnostic features to label missing relationships using as few as $\numlabels$ labeled instances of each relationship.
We learn simple heuristics over these features and assign probabilistic labels to the unlabeled images using a generative model~\cite{ratner2016data,varma2017inferring}. 
We evaluate our method's labeling efficacy using the completely-labeled VRD dataset~\cite{lu2016visual} and find that it achieves an F1 score of $\num{57.66}$, which is $\num{11.84}$ points higher than other standard semi-supervised methods like label propagation~\cite{zhu2002learning}. 
To demonstrate the utility of our generated labels, we train a state-of-the-art scene graph model~\cite{zellers2017neural} (see Figure~\ref{fig:pipeline}) and modify its loss function to support probabilistic labels. 
Our approach achieves \num{47.53} recall@100\footnote{Recall@$K$ is a standard measure for scene graph prediction~\cite{lu2016visual}.} for predicate classification on Visual Genome, improving over the same model trained using only labeled instances by \num{40.97} points.
For scene graph detection, our approach achieves within \num{8.65} recall@100 of the same model trained on the original Visual Genome dataset with $\multiplier\times$ more labeled data. 
We end by comparing our approach to transfer learning, the de-facto choice for learning from limited labels. 
We find that our approach improves by \num{\transferdiff} recall@100 for predicate classification, especially for relationships with high complexity, as it generalizes well to unlabeled subtypes.

Our contributions are three-fold. (1) We introduce the first method to complete visual knowledge bases by finding missing visual relationships (Section~\ref{sec:vrd}). (2) We show the utility of our generated labels in training existing scene graph prediction models (Section~\ref{sec:sgp}). (3) We introduce a metric to characterize the complexity of visual relationships and show it is a strong indicator ($R^2 = \rsquare$) for our semi-supervised method's improvements over transfer learning (Section~\ref{sec:transfer}).
    
    \section{Related work}
    \label{sec:related}
    % Paragraph on NLP KB
\noindent \textbf{Textual knowledge bases} were originally hand-curated by experts to structure facts~\cite{bollacker2008freebase,suchanek2007yago,auer2007dbpedia} (e.g. \relationship{Tokyo}{capital of}{Japan}). 
To scale dataset curation efforts, recent approaches mine knowledge from the web~\cite{carlson2010toward} or hire non-expert annotators to manually curate knowledge~\cite{vrandevcic2014wikidata,bollacker2008freebase}. 
In semi-supervised solutions, a small amount of labeled text is used to extract and exploit patterns in unlabeled sentences~\cite{orbanz2015bayesian,nickel2011three,nickel2012factorizing,nickel2013tensor,hoff2008modeling,anderson1992building}. 
Unfortunately, such approaches cannot be directly applied to visual relationships; textual relations can often be captured by external knowledge or patterns, while visual relationships are often local to an image.  

% Paragraph on relationship detection
\noindent\textbf{Visual relationships} have been studied as spatial priors~\cite{dai2017detecting,galleguillos2008object}, co-occurrences~\cite{yao2010modeling}, language statistics~\cite{yu2017visual,lu2016visual,li2017vip}, and within entity contexts~\cite{li2017scene}. Scene graph prediction models have dealt with the difficulty of learning from incomplete knowledge, as recent methods utilize statistical motifs~\cite{zellers2017neural} or object-relationship dependencies~\cite{xu2017scene,liang2017deep,yang2018graph,zhang2018large}. All these methods limit their inference to the top $50$ most frequently occurring predicate categories and ignore those without enough labeled examples (Figure \ref{fig:sparsity}).

The de-facto solution for limited label problems is \textbf{transfer learning}~\cite{donahue2014decaf, yosinski2014transferable}, which requires that the source domain used for pre-training follows a similar distribution as the target domain. In our setting, the source domain is a dataset of frequently-labeled relationships with thousands of examples~\cite{xu2017scene,liang2017deep,yang2018graph,zhang2018large}, and the target domain is a set of limited label relationships. Despite similar objects in source and target domains, we find that transfer learning has difficulty generalizing to new relationships. 
Our method does not rely on availability of a larger, labeled set of relationships; instead, we use a small labeled set to annotate the unlabeled set of images. 

% Paragraph on weak supervision
To address the issue of gathering enough training labels for machine learning models, \textbf{data programming} has emerged as a popular paradigm. 
This approach learns to model imperfect labeling sources in order to assign training labels to unlabeled data. 
Imperfect labeling sources can come from crowdsourcing~\cite{cheng2015flock}, user-defined heuristics~\cite{bunescu2007learning,shin2015incremental}, multi-instance learning~\cite{hoffmann2011knowledge,riedel2010modeling}, and distant supervision~\cite{craven1999constructing,mintz2009distant}. Often, these imperfect labeling sources take advantage of domain expertise from the user. 
In our case, imperfect labeling sources are automatically generated heuristics, which we aggregate to assign a final probabilistic label to every pair of object proposals.

    \section{Analyzing visual relationships}
    \label{sec:trends}
    
\begin{figure*}[t]
  \centering
  \includegraphics[width=\linewidth]{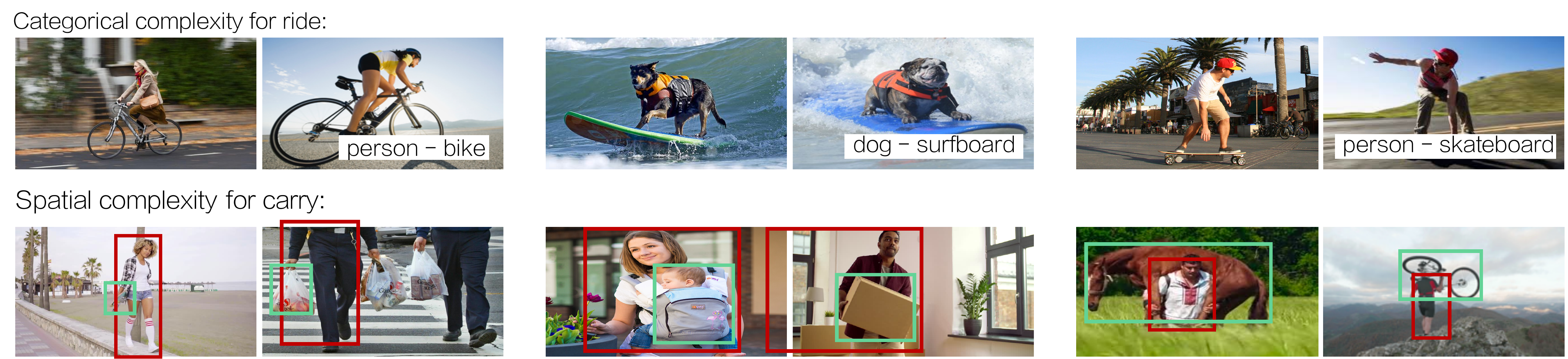}
  \caption{We define the number of subtypes of a relationship as a measure of its complexity. Subtypes can be categorical --- one subtype of \predicate{ride} can be expressed as \relationship{person}{ride}{bike} while another is \relationship{dog}{ride}{surfboard}. Subtypes can also be spatial --- \predicate{carry} has a subtype with a small object carried to the side and another with a large object carried overhead.} 
  \label{fig:variance}
\end{figure*}

\begin{figure*}[t]
  \centering
    \includegraphics[width=0.9\linewidth]{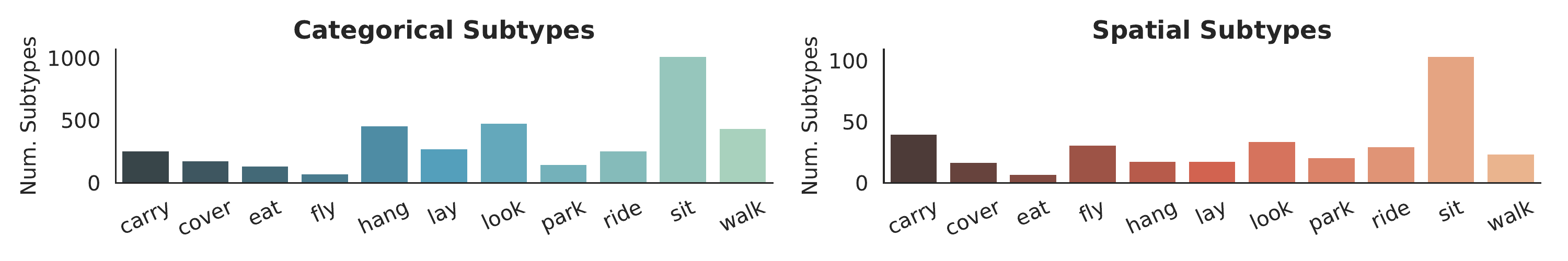}
  \caption{A subset of visual relationships with different levels of complexity as defined by spatial and categorical subtypes. In Section \ref{sec:transfer}, we show how this measure is a good indicator of our semi-supervised method's effectiveness compared to baselines like transfer learning.}
  \label{fig:subtypes}
\end{figure*}

We define the formal terminology used in the rest of the paper and introduce the image-agnostic features that our semi-supervised method relies on. 
Then, we seek quantitative insights into how visual relationships can be described by the properties between its objects.
We ask (1) what image-agnostic features can characterize visual relationships? and (2) given limited labels, how well do our chosen features characterize the complexity of relationships? 
With these in mind, we motivate our model design to generate heuristics that do not overfit to the small amount of labeled data and assign accurate labels to the larger, unlabeled set.

\subsection{Terminology}
A scene graph is a multi-graph $\mathbb{G}$ that consists of objects $o$ as nodes and relationships $r$ as edges. Each object $o_i = \{b_i, c_i\}$ consists of a bounding box $b_i$ and its category $c_i \in \mathbb{C}$ where $\mathbb{C}$ is the set of all possible object categories (e.g. \object{dog}, \object{frisbee}). Relationships are denoted \relationship{subject}{predicate}{object} or \relationship{$o$}{$p$}{$o^\prime$}. $p \in \mathbb{P}$ is a \predicate{predicate}, such as \predicate{ride} and \predicate{eat}. We assume that we have a small labeled set $\{(o, p, o^\prime) \in D_p\}$ of annotated relationships for each predicate $p$. Usually, these datasets are on the order of a $\numlabels$ examples or fewer. For our semi-supervised approach, we also assume that there exists a large set of images $D_U$ without any labeled relationships.

\subsection{Defining image-agnostic features}
It has become common in computer vision to utilize pretrained convolutional neural networks to extract features that represent objects and visual relationships~\cite{xu2017scene,lu2016visual,yang2018graph}. 
Models trained with these features have proven robust in the presence of enough training labels but tend to overfit when presented with limited data (Section~\ref{sec:eval}). Consequently, an open question arises: what other features can we utilize to label relationships with limited data? Previous literature has combined deep learning features with extra information extracted from categorical object labels and relative spatial object locations~\cite{johnson2015image,lu2016visual}. We define categorical features, $<o, -, o^\prime>$, as a concatenation of one-hot vectors of the subject $o$ and object $o^\prime$. We define spatial features as:
\begin{align*}
\frac{x-x^\prime}{w}, \frac{y-y^\prime}{h}, \frac{(y+h)-(y^\prime+h^\prime)}{h}, \\
\frac{(x+w)-(x^\prime+w^\prime)}{w},\frac{h^\prime}{h}, \frac{w^\prime}{w}, \frac{w^\prime h^\prime}{wh}, \frac{w^\prime + h^\prime}{w+h}
\end{align*}
where $b = [y, x, h, w]$ and $b^\prime = [y^\prime, x^\prime, h^\prime, w^\prime]$ are the top-left bounding box coordinates and their widths and heights.

To explore how well spatial and categorical features can describe different visual relationships, we train a simple decision tree model for each relationship. We plot the importances for the top $4$ spatial and categorical features in  Figure~\ref{fig:importance}. Relationships like \predicate{fly} place high importance on the difference in y-coordinate between the subject and object, capturing a characteristic spatial pattern. \predicate{look}, on the other hand, depends on the category of the objects (e.g.~\object{phone}, \object{laptop}, \object{window}) and not on any spatial orientations.

\subsection{Complexity of relationships}
\label{subsec:variance}

To understand the efficacy of image-agnostic features, we'd like to measure how well they can characterize the complexity of particular visual relationships.
As seen in Figure~\ref{fig:variance}, a visual relationship can be defined by a number of image-agnostic features (e.g. a person can \predicate{ride} a bike, or a dog can \predicate{ride} a surfboard).
To systematically define this notion of complexity, we identify \textbf{subtypes} for each visual relationship. 
Each subtype captures one way that a relationship manifests in the dataset. For example, in Figure~\ref{fig:variance}, \predicate{ride} contains one categorical subtype with \relationship{person}{ride}{bike} and another with \relationship{dog}{ride}{surfboard}.
Similarly, a person might \predicate{carry} an object in different relative spatial orientations (e.g. on her head, to her side). 
As shown in Figure~\ref{fig:subtypes}, visual relationships might have significantly different degrees of spatial and categorical complexity, and therefore a different number of subtypes for each. To compute spatial subtypes, we perform mean shift clustering~\cite{cheng1995mean} over the spatial features extracted from all the relationships in Visual Genome. To compute the categorical subtypes, we count the number of unique object categories associated with a relationship.

With access to $\numlabels$ or fewer labeled instances for these visual relationships, it is impossible to capture \textit{all} the subtypes for given relationship and therefore difficult to learn a good representation for the relationship as a whole.
Consequently, we turn to the rules extracted from image-agnostic features and use them to assign labels to the unlabeled data in order to capture a larger proportion of subtypes in each visual relationship.
We posit that this will be advantageous over methods that only use the small labeled set to train a scene graph prediction model, especially for relationships with high complexity, or a large number of subtypes. 
In Section~\ref{sec:transfer}, we find a correlation between our definition of complexity and the performance of our method.

    \section{Approach}
    \label{sec:method}
      
\begin{figure*}[t!]
  \centering
  \includegraphics[width=\linewidth]{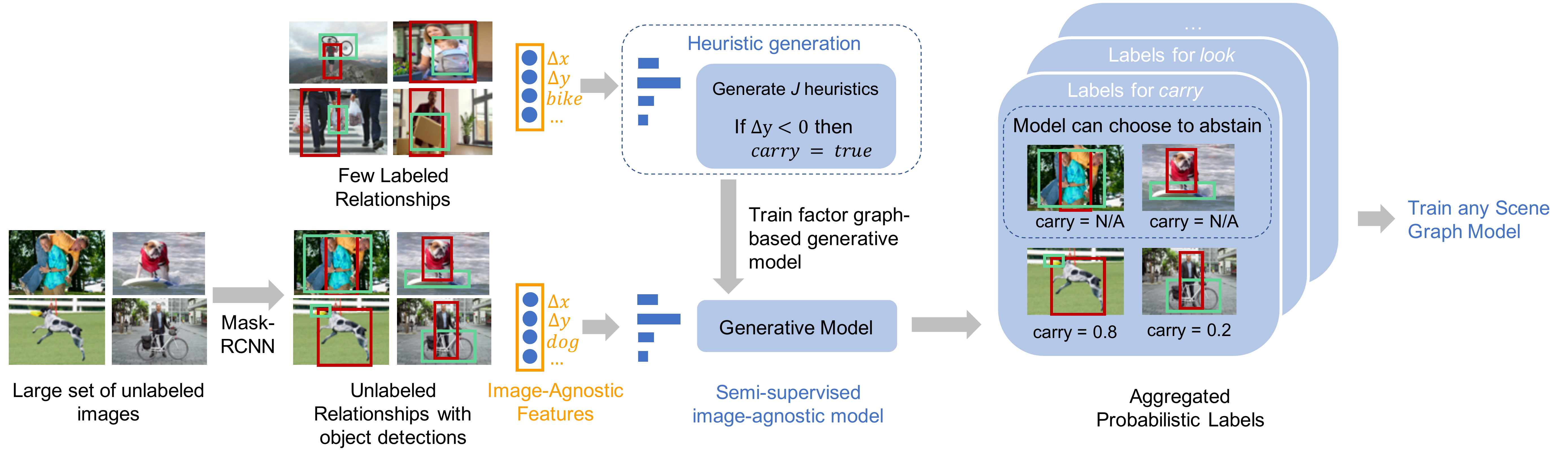}
  \caption{For a relationship (e.g., \predicate{carry}), we use image-agnostic features to automatically create heuristics and then use a generative model to assign probabilistic labels to a large unlabeled set of images. These labels can then be used to train any scene graph prediction model.}
  \label{fig:pipeline}
  \vspace{-1em}
\end{figure*}

We aim to automatically generate labels for missing visual relationships that can be then used to train any downstream scene graph prediction model. We assume that in the long-tail of infrequent relationships, we have a small labeled set $\{(o, p, o^\prime) \in D_p\}$ of annotated relationships for each predicate $p$ (often, on the order of a $\numlabels$ examples or less). As discussed in Section~\ref{sec:trends}, we want to leverage image-agnostic features to learn rules that annotate unlabeled relationships. 

\begin{algorithm}[t]
  \caption{Semi-supervised Alg. to Label Relationships}
  \label{alg:the_only_algorithm_we_have} 
  \begin{algorithmic}[1]
    \scriptsize
    \STATE \textbf{INPUT}: $\{(o, p, o^\prime) \in D_p\} \forall p \in \mathbb{P}$ --- A small dataset of object pairs $(o, o^\prime)$ with multi-class labels for predicates.
    \STATE \textbf{INPUT}: $\{(o, o^\prime)\} \in D_U\}$ --- A large unlabeled dataset of images with objects but no relationship labels.
    \STATE \textbf{INPUT}: $f(\cdot, \cdot)$ --- A function that extracts features from a pair of objects.
    \STATE \textbf{INPUT}: $DT(\cdot)$ --- A decision tree.
    \STATE \textbf{INPUT}: $\textrm{G}(\cdot)$ --- A generative model that assigns probabilistic labels given multiple labels for each datapoint
    \STATE \textbf{INPUT}: $\textrm{train}(\cdot)$ --- Function used to train a scene graph detection model.
    
      \STATE Extract features and labels, $X_p, Y_p := \{f(o, o^\prime), p$ for $(o, p, o^\prime) \in D_{p}\}$, $X_U := \{(f(o, o^\prime)$ for $(o, o^\prime) \in D_{U}\}$

      \STATE Generate heuristics by fitting $J$ decision trees $DT_{fit}(X_p)$ 
      \STATE Assign labels to $(o, o^\prime) \in D_{U}$, $\Lambda = DT_{predict}(X_U)$ for $J$ decision trees. 

      \STATE Learn generative model $\textrm{G}(\Lambda)$ and assign probabilistic labels $\tilde{Y}_U := \textrm{G}(\Lambda)$

    \STATE Train scene graph model, $\textrm{SGM} := \textrm{train}(D_p + D_U, Y_p + \tilde{Y_U})$
    \STATE \textbf{OUTPUT}: SGM($\cdot$)
  \end{algorithmic}
\end{algorithm}

Our approach assigns probabilistic labels to a set $D_U$ of un-annotated images in three steps: (1) we extract image-agnostic features from the objects in the labeled $D_p$ and from the object proposals extracted using an existing object detector~\cite{he2017mask} on unlabeled $D_U$, (2) we generate heuristics over the image-agnostic features, and finally (3) we use a factor-graph based generative model to aggregate and assign probabilistic labels to the unlabeled object pairs in $D_U$. These probabilistic labels, along with $D_p$, are used to train any scene graph prediction model. We describe our approach in Algorithm~\ref{alg:the_only_algorithm_we_have} and show the end-to-end pipeline in Figure~\ref{fig:pipeline}.

\noindent\textbf{Feature extraction: } Our approach uses the image-agnostic features defined in Section~\ref{sec:trends}, which rely on object bounding box and category labels. The features are extracted from ground truth objects in $D_p$ or from object detection outputs in $D_U$ by running existing object detection models~\cite{he2017mask}.

\noindent\textbf{Heuristic generation: } 
We fit decision trees over the labeled relationships' spatial and categorical features to capture image-agnostic rules that define a relationship. These image-agnostic rules are threshold-based conditions that are automatically defined by the decision tree. To limit the complexity of these heuristics and thereby prevent overfitting, we use shallow decision trees~\cite{quinlan1986induction} with different restrictions on depth over each feature set to produce $J$ different decision trees. We then predict labels for the unlabeled set using these heuristics, producing a $\Lambda \in \mathbb{R}^{J \times |D_U|}$ matrix of predictions for the unlabeled relationships. 

Moreover, we only use these heuristics when they have high confidence about their label; we modify $\Lambda$ by converting any predicted label with confidence less than a threshold (empirically chosen to be $2 \times$ random) to an \textit{abstain}, or no label assignment. An example of a heuristic is shown in Figure~\ref{fig:pipeline}: \texttt{if} the subject is above the object, it assigns a positive label for the predicate \predicate{carry}.

\noindent\textbf{Generative model: } 
These heuristics, individually, are noisy and may not assign labels to all object pairs in $D_U$. As a result, we aggregate the labels from all $J$ heuristics.
To do so, we leverage a factor graph-based generative model popular in text-based weak supervision techniques~\cite{xiao2015learning,ratner2016data,takamatsu2012reducing,roth2013combining,alfonseca2012pattern}. This model learns the accuracies of each heuristic to combine their individual labels; the model's output is a probabilistic label for each object pair.

The generative model $G$ uses the following distribution family to relate the latent variable $Y \in \mathbb{R}^{|D_U|}$, the true class, and the labels from the heuristics, $\Lambda$:
$$\pi_{\phi}(\Lambda, Y) = \frac{1}{Z_{\phi}} \exp\left(\phi^T \Lambda Y \right)$$
where $Z_{\phi}$ is a partition function to ensure $\pi$ is normalized. The parameter $\phi \in \mathbb{R}^J$ encodes the average accuracy of each heuristic and is estimated by maximizing the marginal likelihood of the observed heuristic $\Lambda$.
The generative model assigns probabilistic labels by computing $\pi_{\phi}(Y \mid \Lambda(o,o^\prime))$ for each object pair $(o, o^\prime)$ in $D_U$. 

\noindent\textbf{Training scene graph model: } Finally, these probabilistic labels are used to train any scene graph prediction model. While scene graph models are usually trained using a cross-entropy loss~\cite{xu2017scene,lu2016visual,zellers2017neural}, we modify this loss function to take into account errors in the training annotations. We adopt a noise-aware empirical risk minimizer that is often seen in logistic regression as our loss function: 
$$L_{\theta} = \mathbb{E}_{Y \sim \pi}\left[\log\left(1+\exp(-\theta^TV^TY)\right)\right]$$
where $\theta$ is the learned parameters, $\pi$ is the distribution learned by the generative model, $Y$ is the true label, and $V$ are features extracted by any scene graph prediction model.

\begin{table}[t!]
  \centering
    \caption{We validate our approach for labeling missing relationships using only $n=10$ labeled examples by evaluating our probabilistic labels from our semi-supervised approach over the fully-annotated VRD using macro metrics dataset~\cite{lu2016visual}.}
    \label{table:results}
    \small
    \begin{tabular}{@{}lcccc@{}}
    Model ($n=10$) & Prec. & Recall & F1    & Acc. \\ \midrule
    \textsc{Random}      & 5.00      & 5.00   & 5.00  & 5.00     \\
    \textsc{Decision Tree}     & 46.79 & 35.32 & 40.25 &  36.92   \\
    \textsc{Label Propagation}      & \textbf{76.48} & 32.71 & 45.82 & 12.85    \\
    \textsc{Ours (Majority Vote)}      & 55.01 & 57.26 & 56.11  & 40.04    \\
    \textsc{Ours (Categ. + Spat.)}         & 54.83     & \textbf{60.79}  & \textbf{57.66} & \textbf{50.31}    \\
    \end{tabular}
\end{table}
    
    \section{Experiments}
    \label{sec:eval}

\begin{figure*}[t]
  \centering
    \includegraphics[width=\linewidth]{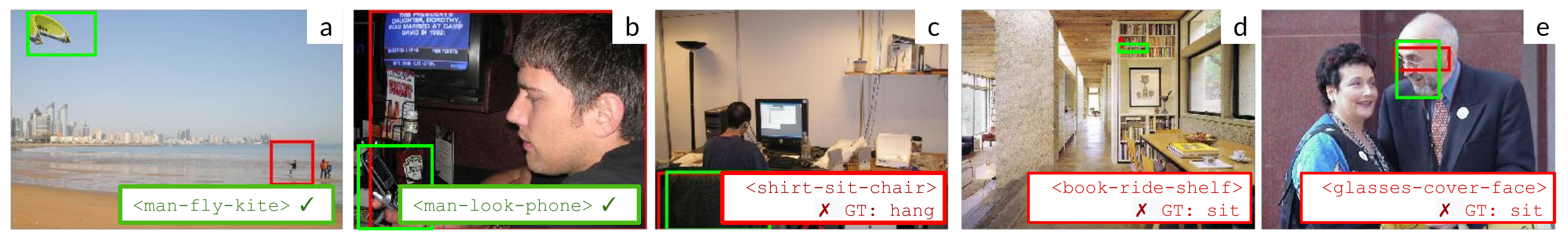}
\caption{(a) Heuristics based on spatial features help predict \relationship{man}{fly}{kite}. (b) Our model learns that \predicate{look} is highly correlated with \object{phone}. (c) We overfit to the importance of \object{chair} as a categorical feature for \predicate{sit}, and fail to identify \predicate{hang} as the correct relationship. (d) We overfit to the spatial positioning associated with \predicate{ride}, where objects are typically longer and directly underneath the subject. (e) Given our image-agnostic features, we produce a reasonable label for \relationship{glass}{cover}{face}. However, our model is incorrect, as two typically different predicates (\predicate{sit} and \predicate{cover}) share a semantic meaning in the context of \relationship{glasses}{?}{face}.}
  \label{fig:qualitative}
    \vspace{-1em}
\end{figure*}
To test our semi-supervised approach for completing visual knowledge bases by annotating missing relationships, we perform a series of experiments and evaluate our framework in several stages. We start by discussing the datasets, baselines, and evaluation metrics used. (1) Our first experiment tests our generative model's ability to find missing relationships in the completely-annotated VRD dataset~\cite{lu2016visual}. (2) Our second experiment demonstrates the utility of our generated labels by using them to train a state-of-the-art scene graph model~\cite{zellers2017neural}. We compare our labels to those from the large Visual Genome dataset~\cite{krishnavisualgenome}. (3) Finally, to show that our semi-supervised method's performance compared to strong baselines in limited label settings, we compare extensively to transfer learning; we focus on a subset of relationships with limited labels, allow the transfer learning model to pretrain on frequent relationships, and demonstrate that our semi-supervised method outperforms transfer learning, which has seen more data. 
Furthermore, we quantify when our method outperforms transfer learning using our metric for measuring relationship complexity (Section~\ref{subsec:variance}).

\begin{table*}[t]
  \centering
  \caption{Results for scene graph prediction tasks with $n=\numlabels$ labeled examples per predicate, reported as recall@K. A state-of-the-art scene graph model trained on labels from our method outperforms those trained with labels generated by other baselines, like transfer learning.} 
  \label{table:superresults}
  \vspace{-0.05em}
  \small
  \begin{tabular}{@{}c@{\hspace{0.4em}} l c@{\hspace{0.2em}} ccc c@{\hspace{0.2em}} ccc c@{\hspace{0.2em}} ccc c@{\hspace{0.2em}} c@{}}
        && \phantom{} & \multicolumn{3}{c}{Scene Graph Detection} &  \phantom{} & \multicolumn{3}{c}{Scene Graph Classification} &  \phantom{} & \multicolumn{3}{c}{Predicate Classification} & \phantom{} \\
      \cmidrule{4-6} \cmidrule{8-10} \cmidrule{12-14} 
  & Model && R$@$20 & R$@$50  & R$@$100 && R$@$20 & R$@$50  & R$@$100  && R$@$20 & R$@$50  & R$@$100  \\
  \midrule
  \multirow{6}{*}{\rotatebox[origin=c]{90}{Baselines}} 
  &\textsc{Baseline [$n = \numlabels$]} & & 0.00 & 0.00 & 0.00 & & 0.04 & 0.04 & 0.04 & & 3.17 & 5.30 & 6.61\\
  &\textsc{Freq} & & 9.01 & 11.01 & 11.64 & & 11.10 & 11.08 & 10.92 & & 20.98 & 20.98 & 20.80\\
  &\textsc{Freq+Overlap} & & 10.16 & 10.84 & 10.86 & & 9.90 & 9.91 & 9.91 & & 20.39 & 20.90 & 22.21\\
  &\textsc{Transfer Learning} & & 11.99 & 14.40 & 16.48 & & 17.10 & 17.91 & 18.16 & & 39.69 & 41.65 & 42.37\\
  &\textsc{Decision tree}~\cite{quinlan1986induction}&  & 11.11  & 12.58  & 13.23 &  & 14.02  & 14.51  & 14.57 &  & 31.75  & 33.02  & 33.35\\
  &\textsc{Label propagation}~\cite{zhu2002learning}&  & 6.48  & 6.74  & 6.83 &  & 9.67  & 9.91  & 9.97 &  & 24.28  & 25.17  & 25.41\\
  \midrule
  \multirow{7}{*}{\rotatebox[origin=c]{90}{Ablations}}
  &\textsc{Ours (Deep)}&  & 2.97  & 3.20  & 3.33 &  & 10.44  & 10.77  & 10.84 &  & 23.16  & 23.93  & 24.17\\
  &\textsc{Ours (Spat.)}&  & 3.26  & 3.20  & 2.91 &  & 10.98  & 11.28  & 11.37 &  & 26.23  & 27.10  & 27.26\\
  &\textsc{Ours (Categ.)}&  & 7.57  & 7.92  & 8.04 &  & 20.83  & \textbf{21.44}  & \textbf{21.57} &  & 43.49  & 44.93  & 45.50\\
  &\textsc{Ours (Categ. + Spat. + Deep)}&  & 7.33  & 7.70  & 7.79 &  & 17.03  & 17.35  & 17.39 &  & 38.90  & 39.87  & 40.02\\
  &\textsc{Ours (Categ. + Spat. + WordVec)}&  & 8.43  & 9.04 & 9.27 &  & 20.39  & 20.90  & 21.21 &  & 45.15  & 46.82  & 47.32\\
  &\textsc{Ours (Majority Vote)}&  & 16.86  & 18.31  & 18.57 &  & 18.96  & 19.57  & 19.66 &  & 44.18  & 45.99  & 46.63\\
  &\textsc{Ours (Categ. + Spat.)} & & \textbf{17.67} & \textbf{18.69} & \textbf{19.28} & & \textbf{20.91} & 21.34 & 21.44 & & \textbf{45.49} & \textbf{47.04} & \textbf{47.53}\\ 
  \midrule
  &\textsc{Oracle [$n_{\text{oracle}} = \multiplier n$]} & & 24.42 & 29.67 & 30.15 & & 30.15 & 30.89 & 31.09 & & 69.23 & 71.40 & 72.15\\
  \end{tabular}
  \end{table*}
  
\begin{figure*}[t]
\centering
      \includegraphics[width=\linewidth]{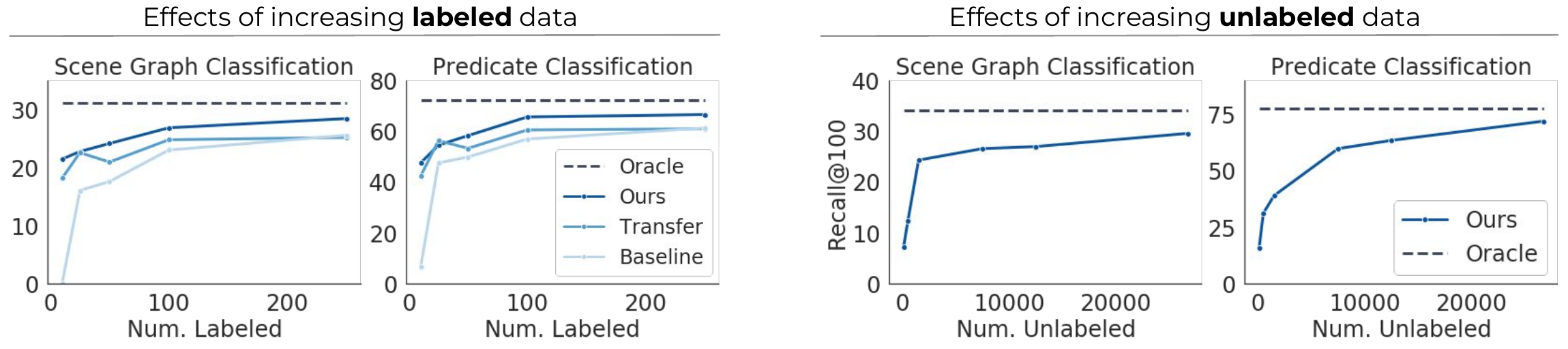}
    \caption{A scene graph model~\cite{zellers2017neural} trained using our labels outperforms both using $\textsc{Transfer Learning}$ labels and using only the $\textsc{Baseline}$ labeled examples consistently across scene graph classification and predicate classification for different amounts of available labeled relationship instances. We also compare to $\textsc{Oracle}$, which is trained with $\multiplier\times$ more labeled data.}
    \label{fig:improvement}
      \vspace{-1em}
  \end{figure*}

\begin{figure*}[t]
    \centering
    \includegraphics[width=0.8\linewidth]{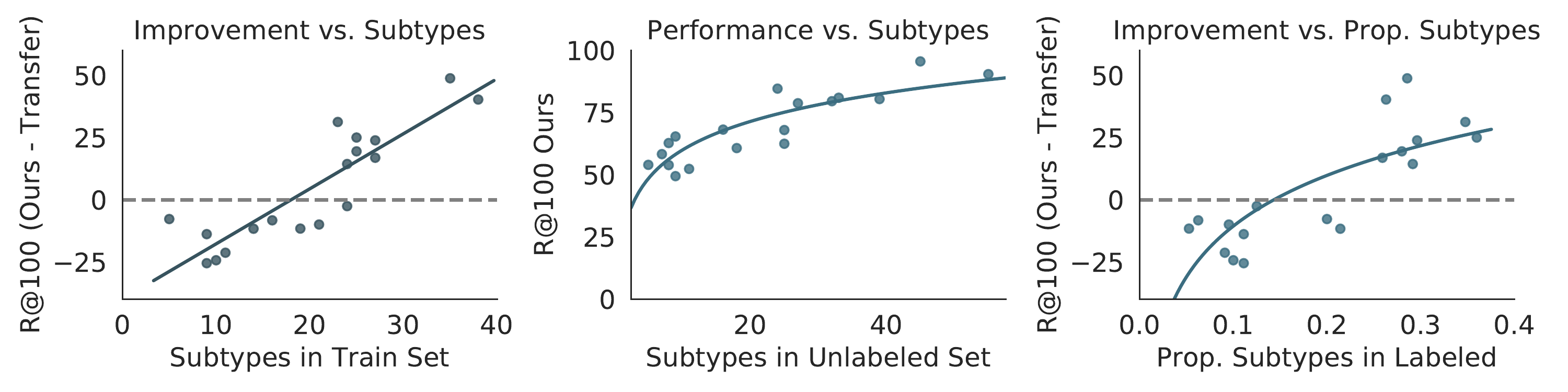}
    \caption{Our method's improvement over transfer learning (in terms of R@100 for predicate classification) is correlated to the number of subtypes in the train set (left), the number of subtypes in the unlabeled set (middle), and the proportion of subtypes in the labeled set (right).}
    \label{fig:trendz}
      \vspace{-1em}
  \end{figure*}

\noindent\textbf{Eliminating synonyms and supersets.} Typically, past scene graph approaches have used $50$ predicates from Visual Genome to study visual relationships. Unfortunately, these $50$ treat synonyms like \predicate{laying on} and \predicate{lying on} as separate classes. To make matters worse, some predicates can be considered a superset of others (i.e. \predicate{above} is a superset of \predicate{riding}). Our method, as well as the baselines, is unable to differentiate between synonyms and supersets. 
For the experiments in this section, we eliminate all supersets and merge all synonyms, resulting in $20$ unique predicates.
In the Appendix (\ref{supp:preds}) we include a list of these predicates and report our method's performance on all $50$ predicates.

\noindent\textbf{Dataset.}
We use two standard datasets, VRD~\cite{lu2016visual} and Visual Genome~\cite{krishnavisualgenome}, to evaluate on tasks related to visual relationships or scene graphs. Each scene graph contains objects localized as bounding boxes in the image along with pairwise relationships connecting them, categorized as action (e.g., \predicate{carry}), possessive (e.g., \predicate{wear}), spatial (e.g., \predicate{above}), or comparative (e.g., \predicate{taller than}) descriptors. Visual Genome is a large visual knowledge base containing $108K$ images. Due to its scale, each scene graph is left with incomplete labels, making it difficult to measure the precision of our semi-supervised algorithm. VRD is a smaller but completely annotated dataset. To show the performance of our semi-supervised method, we measure our method's generated labels on the VRD dataset (Section~\ref{sec:vrd}). Later, we show that the training labels produced can be used to train a large scale scene graph prediction model, evaluated on Visual Genome (Section~\ref{sec:sgp}).

\noindent\textbf{Evaluation metrics.} We measure precision and recall of our generated labels on the VRD dataset's test set (Section~\ref{sec:vrd}). To evaluate a scene graph model trained on our labels, we use three standard evaluation modes for scene graph prediction~\cite{lu2016visual}: (i) scene graph detection (\texttt{SGDET}) which expects input images and predicts bounding box locations, object categories, and predicate labels, (ii) scene graph classification (\texttt{SGCLS}) which expects ground truth boxes and predicts object categories and predicate labels, and (iii) predicate classification (\texttt{PREDCLS}), which expects ground truth bounding boxes and object categories to predict predicate labels. We refer the reader to the paper that introduced these tasks for more details~\cite{lu2016visual}. Finally, we explore how relationship complexity, measured using our definition of subtypes, is correlated with our model's performance relative to transfer learning (Section~\ref{sec:transfer}).

\noindent\textbf{Baselines.}
We compare to alternative methods for generating training labels that can then be used to train downstream scene graph models. $\textsc{oracle}$ is trained on all of Visual Genome, which amounts to $\multiplier\times$ the quantity of labeled relationships in $D_p$; this serves as the upper bound for how well we expect to perform. \textsc{Decision tree}~\cite{quinlan1986induction} fits a single decision tree over the image-agnostic features, learns from labeled examples in $D_p$, and assigns labels to $D_U$. \textsc{Label propagation}~\cite{zhu2002learning} employs a widely-used semi-supervised method and considers the distribution of image-agnostic features in $D_U$ before propagating labels from $D_p$ to $D_U$.

We compare to a strong frequency baselines: (\textsc{Freq}) uses the object counts as priors to make relationship predictions, and \textsc{Freq+Overlap} increments such counts only if the bounding boxes of objects overlap.
We include a \textsc{Transfer Learning} baseline, which is the de-facto choice for training models with limited data~\cite{donahue2014decaf,yosinski2014transferable}.
However, unlike all other methods, transfer learning requires a source dataset to pretrain. 
We treat the source domain as the remaining relationships from the top $50$ in Visual Genome that do not overlap with our chosen relationships. 
We then fine tune with the limited labeled examples for the predicates in $D_p$. 
We note that \textsc{Transfer Learning} has an unfair advantage because there is overlap in objects between its source and target relationship sets. 
Our experiments will show that even with this advantage, our method performs better.

\noindent\textbf{Ablations.} We perform several ablation studies for the image-agnostic features and heuristic aggregation components of our model. 
$\textsc{(Categ.)}$ uses only categorical features, $\textsc{(Spat.)}$ uses only spatial features, $\textsc{(Deep)}$ uses only deep learning features extracted using ResNet50~\cite{He2015} from the union of the object pair's bounding boxes, $\textsc{(Categ. + Spat.)}$ uses both categorical concatenated with spatial features, $\textsc{(Categ. + Spat. + Deep)}$ combines combines all three, and \textsc{Ours (Categ. + Spat. + WordVec)} includes word vectors as richer representations of the categorical features.
\textsc{(Majority Vote)} uses the categorical and spatial features but replaces our generative model with a simple majority voting scheme to aggregate heuristic function outputs.

%=============================================================================
\subsection{Labeling missing relationships}
\label{sec:vrd}
We evaluate our performance in annotating missing relationships in $D_U$. 
Before we use these labels to train scene graph prediction models, we report results comparing our method to baselines in Table~\ref{table:results}. 
On the fully annotated VRD dataset~\cite{lu2016visual}, $\textsc{Ours (Categ. + Spat.)}$ achieves \num{$57.66$} F1 given only $\numlabels$ labeled examples, which is 
\num{$17.41$}, \num{$13.88$}, and \num{$1.55$} 
points better than $\textsc{Label Propagation}$, $\textsc{Decision Tree}$ and $\textsc{Majority Vote}$, respectively.

\noindent\textbf{Qualitative error analysis.} We visualize labels assigned by $\textsc{Ours}$ in Figure~\ref{fig:qualitative} and find that they correspond to image-agnostic rules explored in Figure~\ref{fig:importance}.
In Figure~\ref{fig:qualitative}(a), $\textsc{Ours}$ predicts \predicate{fly} because it learns that \predicate{fly} typically involves objects that have a large difference in y-coordinate. 
In Figure~\ref{fig:qualitative}(b), we correctly label \predicate{look} because \object{phone} is an important categorical feature. 
In some difficult cases, our semi-supervised model fails to generalize beyond the image-agnostic features. In Figure~\ref{fig:qualitative}(c), we mislabel \predicate{hang} as \predicate{sit} by incorrectly relying on the categorical feature \object{chair}, which is one of \predicate{sit}'s important features. 
In Figure~\ref{fig:qualitative}(d), \predicate{ride} typically occurs directly above another object that is slightly larger and assumes \relationship{book}{ride}{shelf} instead of \relationship{book}{sitting on}{shelf}. 
In Figure~\ref{fig:qualitative}(e), our model reasonably classifies \relationship{glasses}{cover}{face}. 
However, \predicate{sit} exhibits the same semantic meaning as \predicate{cover} in this context, and our model incorrectly classifies the example.

%=============================================================================
\subsection{Training Scene graph prediction models}
\label{sec:sgp}
We compare our method's labels to those generated by the baselines described earlier by using them to train three scene graph specific tasks and report results in Table~\ref{table:superresults}. 
We improve over all baselines, including our primary baseline, $\textsc{Transfer Learning}$, by \num{$\transferdiff$} recall@100 for \texttt{PREDCLS}. 
We also achieve within \num{$8.65$} recall@100 of $\textsc{Oracle}$ for \texttt{SGDET}. 
We generate higher quality training labels than $\textsc{Decision Tree}$ and $\textsc{Label Propagation}$, leading to an \num{$13.83$} and \num{$22.12$} recall@100 increase for \texttt{PREDCLS}.

\noindent\textbf{Effect of labeled and unlabeled data.} In Figure~\ref{fig:improvement} (left two graphs), we visualize how \texttt{SGCLS} and \texttt{PREDCLS} performance varies as we reduce the number of labeled examples from $n=250$ to $n=100,50,25,10$. 
We observe greater advantages over $\textsc{Transfer Learning}$ as $n$ decreases, with an increase of \num{$\transferdiff$} recall@100 \texttt{PREDCLS} when $n=10$. This result matches our observations from Section~\ref{sec:trends} because a larger set of labeled examples gives $\textsc{Transfer Learning}$ information about a larger proportion of subtypes for each relationship. In Figure~\ref{fig:improvement} (right two graphs), we visualize our performance as the number of unlabeled data points increase, finding that we approach \textsc{Oracle} performance with more unlabeled examples.

\noindent\textbf{Ablations.} 
$\textsc{Ours (Categ. + Spat. + Deep.)}$ hurts performance by up to \num{$7.51$} recall@100 for \texttt{PREDCLS} because it overfits to image features while $\textsc{Ours (Categ. + Spat.)}$ performs the best. 
We show improvements of \num{$0.71$} recall@100 for \texttt{SGDET} over $\textsc{Ours (MajorityVote)}$, indicating that the generated heuristics indeed have different accuracies and should be weighted differently. 

%=============================================================================
\subsection{Transfer learning vs. semi-supervised learning}
\label{sec:transfer}
Inspired by the recent work comparing transfer learning and semi-supervised learning~\cite{oliver2018realistic}, we characterize when our method is preferred over transfer learning. 
Using the relationship complexity metric based on spatial and categorical subtypes of each predicate (Section~\ref{sec:trends}), we show this trend in Figure~\ref{fig:trendz}. 
When the predicate has a high complexity (as measured by a high number of subtypes), $\textsc{Ours (Categ. + Spat.)}$ outperforms $\textsc{Transfer Learning}$ (Figure~\ref{fig:trendz}, left), with correlation coefficient $R^2=\rsquare$. 
We also evaluate how the number of subtypes in the unlabeled set ($D_U$) affects the performance of our model (Figure~\ref{fig:trendz}, center). 
We find a strong correlation (\num{$R^2 = 0.745$}); our method can effectively assign labels to unlabeled relationships with a large number of subtypes.
We also compare the difference in performance to the proportion of subtypes captured in the labeled set (Figure~\ref{fig:trendz}, right). 
As we hypothesized earlier, $\textsc{Transfer Learning}$ suffers in cases when the labeled set only captures a small portion of the relationship's subtypes. 
This trend (\num{$R^2=0.701$}) explains how $\textsc{Ours (Categ. + Spat.)}$ performs better when given a small portion of labeled subtypes.
    
    \section{Conclusion}
    We introduce the first method that completes visual knowledge bases like Visual Genome by finding missing visual relationships. We define categorical and spatial features as image-agnostic features and introduce a factor-graph based generative model that uses these features to assign probabilistic labels to unlabeled images. 
Our method outperforms baselines in F1 score when finding missing relationships in the complete VRD dataset. Our labels can also be used to train scene graph prediction models with minor modifications to their loss function to accept probabilistic labels. We outperform transfer learning and other baselines and come close to oracle performance of the same model trained on a fraction of labeled data. Finally, we introduce a metric to characterize the complexity of visual relationships and show it is a strong indicator of how our semi-supervised method performs compared to such baselines.

\small
\paragraph{Acknowledgements.} This work was partially funded by the Brown Institute of Media Innovation, the Toyota Research Institute (``TRI''), DARPA under Nos. FA87501720095 and FA86501827865, NIH under No. U54EB020405, NSF under Nos. CCF1763315 and CCF1563078, ONR under No. N000141712266, the Moore Foundation, NXP, Xilinx, LETI-CEA, Intel, Google, NEC, Toshiba, TSMC, ARM, Hitachi, BASF, Accenture, Ericsson, Qualcomm, Analog Devices, the Okawa Foundation, and American Family Insurance, Google Cloud, Swiss Re, NSF Graduate Research Fellowship under No. DGE-114747, 
Joseph W. and Hon Mai Goodman Stanford Graduate Fellowship, 
and members of Stanford DAWN: Intel, Microsoft, Teradata, Facebook, Google, Ant Financial, NEC, SAP, VMWare, and Infosys. The U.S. Government is authorized to reproduce and distribute reprints for Governmental purposes notwithstanding any copyright notation thereon. Any opinions, findings, and conclusions or recommendations expressed in this material are those of the authors and do not necessarily reflect the views, policies, or endorsements, either expressed or implied, of DARPA, NIH, ONR, or the U.S. Government.
\fi

\ifpaper
    {\small
    \bibliographystyle{ieee_fullname}
    \bibliography{main}

\begin{thebibliography}{10}\itemsep=-1pt

\bibitem{alfonseca2012pattern}
Enrique Alfonseca, Katja Filippova, Jean-Yves Delort, and Guillermo Garrido.
\newblock Pattern learning for relation extraction with a hierarchical topic
  model.
\newblock In {\em Proceedings of the 50th Annual Meeting of the Association for
  Computational Linguistics: Short Papers-Volume 2}, pages 54--59. Association
  for Computational Linguistics, 2012.

\bibitem{anderson1992building}
Carolyn~J Anderson, Stanley Wasserman, and Katherine Faust.
\newblock Building stochastic blockmodels.
\newblock {\em Social networks}, 14(1-2):137--161, 1992.

\bibitem{anderson2016spice}
Peter Anderson, Basura Fernando, Mark Johnson, and Stephen Gould.
\newblock Spice: Semantic propositional image caption evaluation.
\newblock In {\em European Conference on Computer Vision}, pages 382--398.
  Springer, 2016.

\bibitem{auer2007dbpedia}
S{\"o}ren Auer, Christian Bizer, Georgi Kobilarov, Jens Lehmann, Richard
  Cyganiak, and Zachary Ives.
\newblock Dbpedia: A nucleus for a web of open data.
\newblock In {\em The semantic web}, pages 722--735. Springer, 2007.

\bibitem{bollacker2008freebase}
Kurt Bollacker, Colin Evans, Praveen Paritosh, Tim Sturge, and Jamie Taylor.
\newblock Freebase: a collaboratively created graph database for structuring
  human knowledge.
\newblock In {\em Proceedings of the 2008 ACM SIGMOD international conference
  on Management of data}, pages 1247--1250. AcM, 2008.

\bibitem{bordes2014semantic}
Antoine Bordes, Xavier Glorot, Jason Weston, and Yoshua Bengio.
\newblock A semantic matching energy function for learning with
  multi-relational data.
\newblock {\em Machine Learning}, 94(2):233--259, 2014.

\bibitem{bordes2013translating}
Antoine Bordes, Nicolas Usunier, Alberto Garcia-Duran, Jason Weston, and Oksana
  Yakhnenko.
\newblock Translating embeddings for modeling multi-relational data.
\newblock In {\em Advances in neural information processing systems}, pages
  2787--2795, 2013.

\bibitem{bunescu2007learning}
Razvan Bunescu and Raymond Mooney.
\newblock Learning to extract relations from the web using minimal supervision.
\newblock In {\em Proceedings of the 45th Annual Meeting of the Association of
  Computational Linguistics}, pages 576--583, 2007.

\bibitem{carlson2010toward}
Andrew Carlson, Justin Betteridge, Bryan Kisiel, Burr Settles, Estevam~R
  Hruschka~Jr, and Tom~M Mitchell.
\newblock Toward an architecture for never-ending language learning.
\newblock In {\em AAAI}, volume~5, page~3. Atlanta, 2010.

\bibitem{cheng2015flock}
Justin Cheng and Michael~S Bernstein.
\newblock Flock: Hybrid crowd-machine learning classifiers.
\newblock In {\em Proceedings of the 18th ACM conference on computer supported
  cooperative work \& social computing}, pages 600--611. ACM, 2015.

\bibitem{cheng1995mean}
Yizong Cheng.
\newblock Mean shift, mode seeking, and clustering.
\newblock {\em IEEE transactions on pattern analysis and machine intelligence},
  17(8):790--799, 1995.

\bibitem{craven1999constructing}
Mark Craven, Johan Kumlien, et~al.
\newblock Constructing biological knowledge bases by extracting information
  from text sources.
\newblock In {\em ISMB}, volume 1999, pages 77--86, 1999.

\bibitem{culotta2004dependency}
Aron Culotta and Jeffrey Sorensen.
\newblock Dependency tree kernels for relation extraction.
\newblock In {\em Proceedings of the 42nd annual meeting on association for
  computational linguistics}, page 423. Association for Computational
  Linguistics, 2004.

\bibitem{dai2017detecting}
Bo Dai, Yuqi Zhang, and Dahua Lin.
\newblock Detecting visual relationships with deep relational networks.
\newblock In {\em 2017 IEEE Conference on Computer Vision and Pattern
  Recognition (CVPR)}, pages 3298--3308. IEEE, 2017.

\bibitem{donahue2014decaf}
Jeff Donahue, Yangqing Jia, Oriol Vinyals, Judy Hoffman, Ning Zhang, Eric
  Tzeng, and Trevor Darrell.
\newblock Decaf: A deep convolutional activation feature for generic visual
  recognition.
\newblock In {\em International conference on machine learning}, pages
  647--655, 2014.

\bibitem{galleguillos2008object}
Carolina Galleguillos, Andrew Rabinovich, and Serge Belongie.
\newblock Object categorization using co-occurrence, location and appearance.
\newblock In {\em Computer Vision and Pattern Recognition, 2008. CVPR 2008.
  IEEE Conference on}, pages 1--8. IEEE, 2008.

\bibitem{gardner2014incorporating}
Matt Gardner, Partha Talukdar, Jayant Krishnamurthy, and Tom Mitchell.
\newblock Incorporating vector space similarity in random walk inference over
  knowledge bases.
\newblock In {\em Proceedings of the 2014 Conference on Empirical Methods in
  Natural Language Processing (EMNLP)}, pages 397--406, 2014.

\bibitem{guodong2005exploring}
Zhou GuoDong, Su Jian, Zhang Jie, and Zhang Min.
\newblock Exploring various knowledge in relation extraction.
\newblock In {\em Proceedings of the 43rd annual meeting on association for
  computational linguistics}, pages 427--434. Association for Computational
  Linguistics, 2005.

\bibitem{he2017mask}
Kaiming He, Georgia Gkioxari, Piotr Doll{\'a}r, and Ross Girshick.
\newblock Mask r-cnn.
\newblock In {\em Computer Vision (ICCV), 2017 IEEE International Conference
  on}, pages 2980--2988. IEEE, 2017.

\bibitem{He2015}
Kaiming He, Xiangyu Zhang, Shaoqing Ren, and Jian Sun.
\newblock Deep residual learning for image recognition.
\newblock {\em arXiv preprint arXiv:1512.03385}, 2015.

\bibitem{hoff2008modeling}
Peter Hoff.
\newblock Modeling homophily and stochastic equivalence in symmetric relational
  data.
\newblock In {\em Advances in neural information processing systems}, pages
  657--664, 2008.

\bibitem{hoffmann2011knowledge}
Raphael Hoffmann, Congle Zhang, Xiao Ling, Luke Zettlemoyer, and Daniel~S Weld.
\newblock Knowledge-based weak supervision for information extraction of
  overlapping relations.
\newblock In {\em Proceedings of the 49th Annual Meeting of the Association for
  Computational Linguistics: Human Language Technologies-Volume 1}, pages
  541--550. Association for Computational Linguistics, 2011.

\bibitem{johnson2018image}
Justin Johnson, Agrim Gupta, and Li Fei-Fei.
\newblock Image generation from scene graphs.
\newblock {\em arXiv preprint arXiv:1804.01622}, 2018.

\bibitem{johnson2017inferring}
Justin Johnson, Bharath Hariharan, Laurens van~der Maaten, Judy Hoffman, Li
  Fei-Fei, C~Lawrence Zitnick, and Ross Girshick.
\newblock Inferring and executing programs for visual reasoning.
\newblock {\em arXiv preprint arXiv:1705.03633}, 2017.

\bibitem{johnson2015image}
Justin Johnson, Ranjay Krishna, Michael Stark, Li-Jia Li, David Shamma, Michael
  Bernstein, and Li Fei-Fei.
\newblock Image retrieval using scene graphs.
\newblock In {\em Proceedings of the IEEE conference on computer vision and
  pattern recognition}, pages 3668--3678, 2015.

\bibitem{krishna2018referring}
Ranjay Krishna, Ines Chami, Michael Bernstein, and Li Fei-Fei.
\newblock Referring relationships.
\newblock In {\em IEEE Conference on Computer Vision and Pattern Recognition},
  2018.

\bibitem{krishnavisualgenome}
Ranjay Krishna, Yuke Zhu, Oliver Groth, Justin Johnson, Kenji Hata, Joshua
  Kravitz, Stephanie Chen, Yannis Kalantidis, Li-Jia Li, David~A Shamma, et~al.
\newblock Visual genome: Connecting language and vision using crowdsourced
  dense image annotations.
\newblock {\em International Journal of Computer Vision}, 123(1):32--73, 2017.

\bibitem{li2017vip}
Yikang Li, Wanli Ouyang, Xiaogang Wang, and Xiao'Ou Tang.
\newblock Vip-cnn: Visual phrase guided convolutional neural network.
\newblock In {\em Computer Vision and Pattern Recognition (CVPR), 2017 IEEE
  Conference on}, pages 7244--7253. IEEE, 2017.

\bibitem{li2017scene}
Yikang Li, Wanli Ouyang, Bolei Zhou, Kun Wang, and Xiaogang Wang.
\newblock Scene graph generation from objects, phrases and region captions.
\newblock In {\em Proceedings of the IEEE Conference on Computer Vision and
  Pattern Recognition}, pages 1261--1270, 2017.

\bibitem{liang2017deep}
Xiaodan Liang, Lisa Lee, and Eric~P Xing.
\newblock Deep variation-structured reinforcement learning for visual
  relationship and attribute detection.
\newblock In {\em Computer Vision and Pattern Recognition (CVPR), 2017 IEEE
  Conference on}, pages 4408--4417. IEEE, 2017.

\bibitem{lu2016visual}
Cewu Lu, Ranjay Krishna, Michael Bernstein, and Li Fei-Fei.
\newblock Visual relationship detection with language priors.
\newblock In {\em European Conference on Computer Vision}, pages 852--869.
  Springer, 2016.

\bibitem{mintz2009distant}
Mike Mintz, Steven Bills, Rion Snow, and Dan Jurafsky.
\newblock Distant supervision for relation extraction without labeled data.
\newblock In {\em Proceedings of the Joint Conference of the 47th Annual
  Meeting of the ACL and the 4th International Joint Conference on Natural
  Language Processing of the AFNLP: Volume 2-Volume 2}, pages 1003--1011.
  Association for Computational Linguistics, 2009.

\bibitem{nickel2013tensor}
Maximilian Nickel.
\newblock {\em Tensor factorization for relational learning}.
\newblock PhD thesis, lmu, 2013.

\bibitem{nickel2011three}
Maximilian Nickel, Volker Tresp, and Hans-Peter Kriegel.
\newblock A three-way model for collective learning on multi-relational data.
\newblock In {\em ICML}, volume~11, pages 809--816, 2011.

\bibitem{nickel2012factorizing}
Maximilian Nickel, Volker Tresp, and Hans-Peter Kriegel.
\newblock Factorizing yago: scalable machine learning for linked data.
\newblock In {\em Proceedings of the 21st international conference on World
  Wide Web}, pages 271--280. ACM, 2012.

\bibitem{oliver2018realistic}
Avital Oliver, Augustus Odena, Colin Raffel, Ekin~D Cubuk, and Ian~J
  Goodfellow.
\newblock Realistic evaluation of deep semi-supervised learning algorithms.
\newblock {\em arXiv preprint arXiv:1804.09170}, 2018.

\bibitem{orbanz2015bayesian}
Peter Orbanz and Daniel~M Roy.
\newblock Bayesian models of graphs, arrays and other exchangeable random
  structures.
\newblock {\em IEEE transactions on pattern analysis and machine intelligence},
  37(2):437--461, 2015.

\bibitem{quinlan1986induction}
J.~Ross Quinlan.
\newblock Induction of decision trees.
\newblock {\em Machine learning}, 1(1):81--106, 1986.

\bibitem{ratner2016data}
Alexander~J Ratner, Christopher~M De~Sa, Sen Wu, Daniel Selsam, and Christopher
  R\'{e}.
\newblock Data programming: Creating large training sets, quickly.
\newblock In D.~D. Lee, M. Sugiyama, U.~V. Luxburg, I. Guyon, and R. Garnett,
  editors, {\em Advances in Neural Information Processing Systems 29}, pages
  3567--3575. Curran Associates, Inc., 2016.

\bibitem{riedel2010modeling}
Sebastian Riedel, Limin Yao, and Andrew McCallum.
\newblock Modeling relations and their mentions without labeled text.
\newblock In {\em Joint European Conference on Machine Learning and Knowledge
  Discovery in Databases}, pages 148--163. Springer, 2010.

\bibitem{roth2013combining}
Benjamin Roth and Dietrich Klakow.
\newblock Combining generative and discriminative model scores for distant
  supervision.
\newblock In {\em Proceedings of the 2013 Conference on Empirical Methods in
  Natural Language Processing}, pages 24--29, 2013.

\bibitem{schuster2015generating}
Sebastian Schuster, Ranjay Krishna, Angel Chang, Li Fei-Fei, and Christopher~D
  Manning.
\newblock Generating semantically precise scene graphs from textual
  descriptions for improved image retrieval.
\newblock In {\em Proceedings of the fourth workshop on vision and language},
  pages 70--80, 2015.

\bibitem{shin2015incremental}
Jaeho Shin, Sen Wu, Feiran Wang, Christopher De~Sa, Ce Zhang, and Christopher
  R{\'e}.
\newblock Incremental knowledge base construction using deepdive.
\newblock {\em Proceedings of the VLDB Endowment}, 8(11):1310--1321, 2015.

\bibitem{suchanek2007yago}
Fabian~M Suchanek, Gjergji Kasneci, and Gerhard Weikum.
\newblock Yago: a core of semantic knowledge.
\newblock In {\em Proceedings of the 16th international conference on World
  Wide Web}, pages 697--706. ACM, 2007.

\bibitem{takamatsu2012reducing}
Shingo Takamatsu, Issei Sato, and Hiroshi Nakagawa.
\newblock Reducing wrong labels in distant supervision for relation extraction.
\newblock In {\em Proceedings of the 50th Annual Meeting of the Association for
  Computational Linguistics: Long Papers-Volume 1}, pages 721--729. Association
  for Computational Linguistics, 2012.

\bibitem{varma2017inferring}
Paroma Varma, Bryan~D He, Payal Bajaj, Nishith Khandwala, Imon Banerjee, Daniel
  Rubin, and Christopher R{\'e}.
\newblock Inferring generative model structure with static analysis.
\newblock In {\em Advances in Neural Information Processing Systems}, pages
  239--249, 2017.

\bibitem{vrandevcic2014wikidata}
Denny Vrande{\v{c}}i{\'c} and Markus Kr{\"o}tzsch.
\newblock Wikidata: a free collaborative knowledgebase.
\newblock {\em Communications of the ACM}, 57(10):78--85, 2014.

\bibitem{xiao2015learning}
Tong Xiao, Tian Xia, Yi Yang, Chang Huang, and Xiaogang Wang.
\newblock Learning from massive noisy labeled data for image classification.
\newblock In {\em Proceedings of the IEEE Conference on Computer Vision and
  Pattern Recognition}, pages 2691--2699, 2015.

\bibitem{xu2017scene}
Danfei Xu, Yuke Zhu, Christopher~B Choy, and Li Fei-Fei.
\newblock Scene graph generation by iterative message passing.
\newblock In {\em Proceedings of the IEEE Conference on Computer Vision and
  Pattern Recognition}, volume~2, 2017.

\bibitem{yang2018graph}
Jianwei Yang, Jiasen Lu, Stefan Lee, Dhruv Batra, and Devi Parikh.
\newblock Graph r-cnn for scene graph generation.
\newblock {\em arXiv preprint arXiv:1808.00191}, 2018.

\bibitem{yao2010modeling}
Bangpeng Yao and Li Fei-Fei.
\newblock Modeling mutual context of object and human pose in human-object
  interaction activities.
\newblock In {\em Computer Vision and Pattern Recognition (CVPR), 2010 IEEE
  Conference on}, pages 17--24. IEEE, 2010.

\bibitem{yosinski2014transferable}
Jason Yosinski, Jeff Clune, Yoshua Bengio, and Hod Lipson.
\newblock How transferable are features in deep neural networks?
\newblock In {\em Advances in neural information processing systems}, pages
  3320--3328, 2014.

\bibitem{yu2017visual}
Ruichi Yu, Ang Li, Vlad~I Morariu, and Larry~S Davis.
\newblock Visual relationship detection with internal and external linguistic
  knowledge distillation.
\newblock {\em arXiv preprint arXiv:1707.09423}, 2017.

\bibitem{zellers2017neural}
Rowan Zellers, Mark Yatskar, Sam Thomson, and Yejin Choi.
\newblock Neural motifs: Scene graph parsing with global context.
\newblock {\em arXiv preprint arXiv:1711.06640}, 2017.

\bibitem{zhang2018large}
Ji Zhang, Yannis Kalantidis, Marcus Rohrbach, Manohar Paluri, Ahmed Elgammal,
  and Mohamed Elhoseiny.
\newblock Large-scale visual relationship understanding.
\newblock {\em arXiv preprint arXiv:1804.10660}, 2018.

\bibitem{zhou2007tree}
Guodong Zhou, Min Zhang, DongHong Ji, and Qiaoming Zhu.
\newblock Tree kernel-based relation extraction with context-sensitive
  structured parse tree information.
\newblock In {\em Proceedings of the 2007 Joint Conference on Empirical Methods
  in Natural Language Processing and Computational Natural Language Learning
  (EMNLP-CoNLL)}, 2007.

\bibitem{zhu2002learning}
Xiaojin Zhu and Zoubin Ghahramani.
\newblock Learning from labeled and unlabeled data with label propagation.
\newblock {\em Technical Report}, 2002.

\end{thebibliography}
    }
\fi

\ifsupplementary
    \clearpage
    \renewcommand{\thesection}{A\arabic{section}}
    \setcounter{section}{0} % Restart numbering in Appendix
    \renewcommand{\thetable}{A\arabic{table}}
    \setcounter{table}{0} % Restart numbering in Appendix
    \renewcommand{\thefigure}{A\arabic{figure}}
    \setcounter{figure}{0} % Restart numbering in Appendix
    \section{Appendix}
    
\begin{figure*}[ht!]
  \centering
  \includegraphics[width=\linewidth]{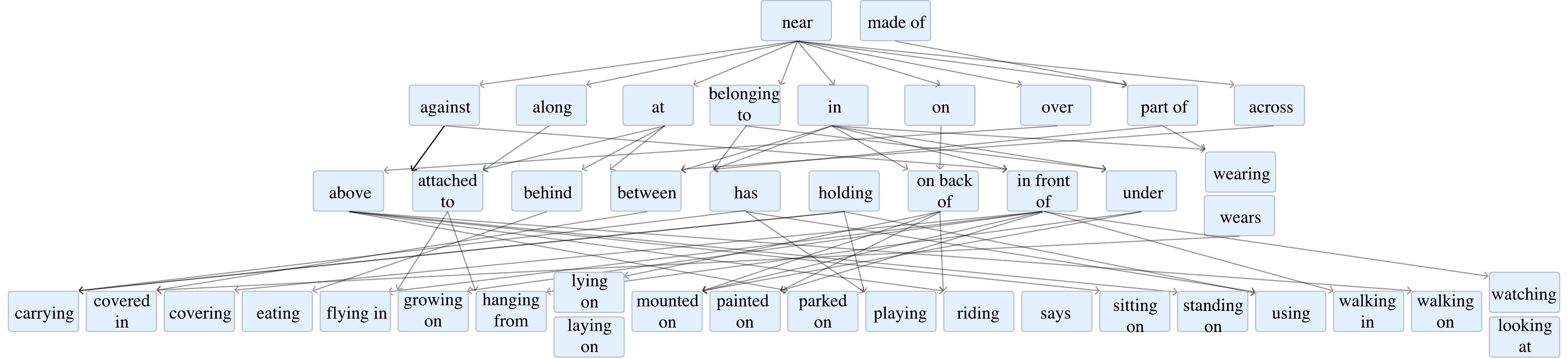}
  % Modify graph here: https://www.figma.com/file/V24rxVMLmN07Q3TepNmIiPmL/ICCV_supersets?node-id=0%3A1
  \caption{We define dependencies between predicates to determine which ones to include in the evaluation of our method. Directional arrows indicate supersets, and stacked nodes indicates synonyms. \textit{Note:} \texttt{says} has no parents, so we treat this as a leaf node in our experimental setup.}
  \label{fig:supersets}
\end{figure*}

\begin{table*}[ht!]
  \centering
  \label{table:50results}
  \caption{Results for top 50 predicates in Visual Genome.}
  \small
  %\ra{1.2}
  \begin{tabular}{@{}c@{\hspace{0.4em}} l c@{\hspace{0.2em}} ccc c@{\hspace{0.2em}} ccc c@{\hspace{0.2em}} ccc c@{\hspace{0.2em}} c@{}}
        && \phantom{} & \multicolumn{3}{c}{Scene Graph Detection} &  \phantom{} & \multicolumn{3}{c}{Scene Graph Classification} &  \phantom{} & \multicolumn{3}{c}{Predicate Classification} & \phantom{} \\
      \cmidrule{4-6} \cmidrule{8-10} \cmidrule{12-14} %\my89{--blame->\cmidrule{10}<--blame-- this change made it not compile and crash the project. whhhhyyy?}
  & Model && R$@$20 & R$@$50  & R$@$100 && R$@$20 & R$@$50  & R$@$100  && R$@$20 & R$@$50  & R$@$100  \\
  \midrule
  &\textsc{Baseline [$n = \numlabels$]} & & 1.06 & 1.80 & 2.66 & & 4.70 & 6.00 & 5.43 && 9.63 & 12.17 & 13.07\\
  &\textsc{Ours (Categ. + Spat.)} & & \textbf{4.04} & \textbf{6.75} & \textbf{8.64} & & \textbf{12.69} & \textbf{13.91} & \textbf{14.16} & & \textbf{24.72} & \textbf{27.76} & \textbf{28.53}\\ 
  \midrule
  &\textsc{Oracle [$n_{\text{oracle}} = 44 n$]} & & 14.20 & 20.61 & 25.44 & & 33.58 & 35.52 & 35.92 & & 62.00 & 66.92 & 68.02\\
  \end{tabular}
\end{table*}

\label{supp:preds}
\subsection{Choice of predicates}
As discussed in Section~$5$, we used a subset of predicates for our primary experiments because the full 50 predicates represent a large number of synonyms and supersets for each predicate. We identified these dependencies between predicates as a directed graph, and selected the leaf nodes (bottom row) as our chosen predicates in Figure~\ref{fig:supersets}.

\subsection{Performance on all $50$ predicates}
Furthermore, we have included results on the full set of 50 predicates in Table~\ref{table:50results}. Note that we are unable to evaluate against our primary baseline, transfer learning, because we have utilized all potential source domain predicates in this experiment. We see that our method improves over the baseline approach using $n=10$ labeled examples per relationship by $15.46$ R@100 for \texttt{PREDCLS}. We see similar trends across the various ablations of our model and therefore, only report the our best model.
\fi

\end{document}